\documentclass[letterpaper, 10 pt, conference]{ieeeconf}  

\IEEEoverridecommandlockouts                              

\usepackage{graphics,caption} 
\graphicspath{ {figure/} }
\usepackage{graphicx} 
\usepackage{epsfig} 
\usepackage{times} 
\usepackage{amsmath} 
\usepackage{amssymb} 
\usepackage{cite}
\usepackage{booktabs}
\usepackage{threeparttable}
\usepackage{multirow}
\if CLASSOPTIONcompsoc
\usepackage[caption=false, font=normalsize, labelfont=sf, textfont=sf]{subfig}
\else
\usepackage[caption=false, font=footnotesize]{subfig}
\usepackage{url}
\usepackage{mathrsfs}
\usepackage[colorlinks,linkcolor=blue]{hyperref}

\usepackage{float}
\usepackage{stfloats}

\usepackage{tcolorbox}
\tcbuselibrary{skins}
\usepackage{lipsum}

\title{\LARGE \bf
Co-NavGPT: Multi-Robot Cooperative Visual Semantic Navigation  \\
Using Vision Language Models
}

\author{Bangguo Yu, Qihao Yuan, Kailai Li, Hamidreza Kasaei, and Ming Cao
\thanks{All authors are with the Faculty of Science and Engineering, University of Groningen, 9747 AG Groningen, the Netherlands. {\tt\small \{b.yu, qihao.yuan, kailai.li, hamidreza.kasaei, m.cao\}@rug.nl}}%
}

\begin{document}

\maketitle
\thispagestyle{empty}
\pagestyle{empty}

\begin{abstract}

Visual target navigation is a critical capability for autonomous robots operating in unknown environments, particularly in human-robot interaction scenarios. While classical and learning-based methods have shown promise, most existing approaches lack common-sense reasoning and are typically designed for single-robot settings, leading to reduced efficiency and robustness in complex environments.
To address these limitations, we introduce Co-NavGPT, a novel framework that integrates a Vision Language Model (VLM) as a global planner to enable common-sense multi-robot visual target navigation. Co-NavGPT aggregates sub-maps from multiple robots with diverse viewpoints into a unified global map, encoding robot states and frontier regions. The VLM uses this information to assign frontiers across the robots, facilitating coordinated and efficient exploration.
Experiments on the Habitat-Matterport 3D (HM3D) demonstrate that Co-NavGPT outperforms existing baselines in terms of success rate and navigation efficiency, without requiring task-specific training. Ablation studies further confirm the importance of semantic priors from the VLM. We also validate the framework in real-world scenarios using quadrupedal robots.
Supplementary video and code are available at: \href{https://sites.google.com/view/co-navgpt2}{https://sites.google.com/view/co-navgpt2}.

\end{abstract}

\section{INTRODUCTION}



Humans efficiently explore complex environments by leveraging common-sense knowledge of environmental structures and collaborative strategies. Similarly, autonomous robots require reasoning abilities to effectively navigate and explore their environments. For example, consider a person instructing two robots in his house, ``Hey, robots, please bring my cell phone to me.'' In this task, implicitly, the first critical step is for the two robots to locate the specific object, namely the cell phone, collaboratively. Although substantial progress has been made in object-goal navigation and multi-robot exploration, efficiently enabling robots to collaboratively identify and locate target objects from visual inputs remains challenging due to semantic complexity and environmental intricacies.
In this paper, we address the task of multi-robot visual target navigation, wherein multiple robots collaboratively explore unknown environments to efficiently locate a specific target object. This capability serves as the enabling functionality for various practical robotic applications.

\begin{figure}[t]
	\centering
	\includegraphics[scale=0.6]{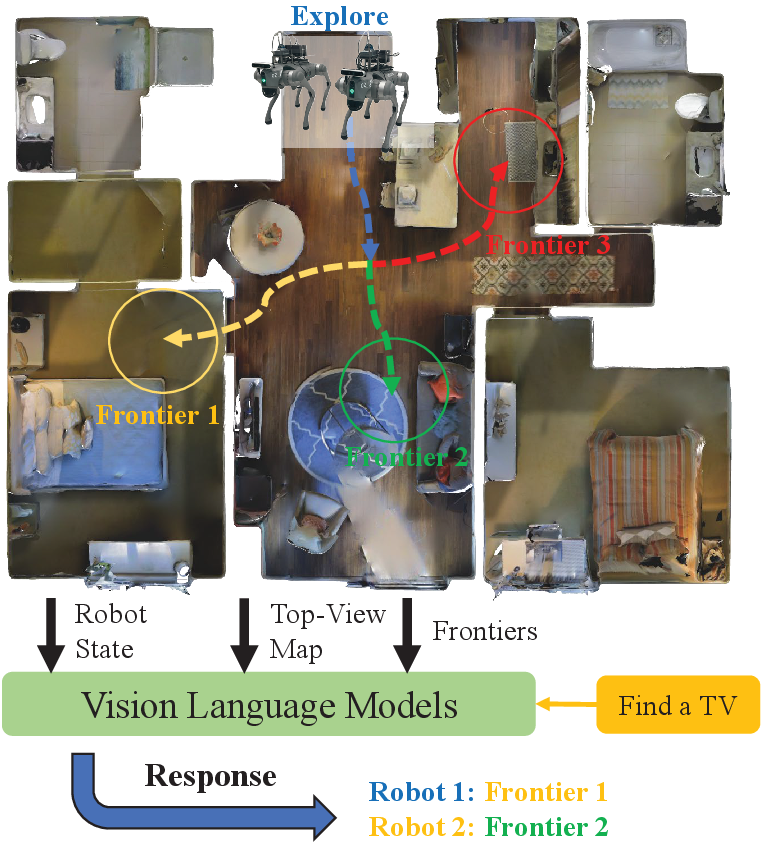}
	\caption{Two robots visual target navigation example. When multiple unexplored frontiers are detected, the vision language model assigns the frontier for each robot based on the current observation and the target object.}
	\label{fig:demo}
\end{figure}

Recent advances in simulation platforms~\cite{aihabitat,Gibson}, large-scale 3D scene datasets~\cite{Ramakrishnan2021a,Gibson}, map-based representations~\cite{vlmaps, Conceptgraph}, and large foundation models~\cite{Khandelwal2022, openai2024gpt4o} have spurred significant interest in visual target navigation. Existing approaches can be broadly categorized into end-to-end frameworks~\cite{Zhu2017, Khandelwal2022} and modular pipelines~\cite{Chaplot2020b, poni}, which aim to improve navigation performance by incorporating spatial and semantic features~\cite{Lyu2022, Druon2020} through supervised or reinforcement learning.
More recently, Vision Language Models (VLMs) have been introduced into navigation pipelines~\cite{Liang2022, Zhou2024}, leveraging their strengths in semantic reasoning and object grounding. VLMs provide high-level semantic priors, enabling better understanding of object relationships and scene structure.
However, the vast majority of existing methods are designed for single-robot settings. In large or complex environments, single-agent navigation often suffers from poor efficiency due to the need to explore vast unknown areas. Additionally, such systems are vulnerable to failure, as a single incorrect decision or unexpected obstacle can significantly delay or derail the navigation process. These limitations motivate the need for multi-robot collaboration in visual target navigation.

Learning-based approaches typically require large amounts of data and corresponding computation to generalize across diverse environments, a challenge that becomes more profound in multi-robot scenarios~\cite{Liu2022}. With the emergence of VLMs, new opportunities have arisen for zero-shot and few-shot planning in embodied tasks. VLMs possess inherent world knowledge and have shown promise as high-level planners in instruction-following tasks~\cite{Huang2022b}, as well as in long-horizon, multi-agent coordination~\cite{Song2022,Hong2023,Zhang2023}. These capabilities make VLMs an ideal candidate for enabling common-sense reasoning in multi-robot exploration and navigation tasks.

This paper addresses the challenge of multi-robot visual semantic navigation, wherein multiple robots collaboratively locate a target object in unknown environments. Specifically, we introduce Co-NavGPT, a novel framework that leverages vision-language models to generate efficient exploration and search policies for multi-robot cooperation. Within this framework, a VLM functions as a global planner, strategically assigning unexplored frontier regions to each robot. An illustrative example of visual target navigation, locating a television, is presented in Fig. \ref*{fig:demo}. After mapping their environment, robots must select their next exploration frontier. Utilizing contextual information from the observed environment map, each robot's state, and the given navigation target, a VLM allocates the most relevant frontiers to individual robots, enhancing collaborative search efficiency.
We evaluate Co-NavGPT on the Habitat simulation platform \cite{Savva2019}, comparing its navigation performance against existing multi-robot methods within extensive photorealistic 3D environments from HM3D \cite{Ramakrishnan2021a}. Ablation experiments further validate the effectiveness of integrating VLMs in multi-robot navigation tasks. Additionally, we demonstrate our method's practical applicability using two quadrupedal robots in real-world experiments. In contrast to other multi-robot navigation methods, Co-NavGPT uniquely utilizes VLMs to encode frontier-enhanced environment representations, thereby significantly improving scene understanding and cooperative navigation efficiency. Our results highlight the substantial potential of VLMs for managing complex multi-robot collaborative tasks.

Our contributions are summarized as follows:

\begin{itemize}

    \item We propose Co-NavGPT,  a novel framework that merges multi-robot observations into a global semantic map. The framework uses vision-language models to guide robots toward efficient collaborative exploration and navigation in unknown environments.
    \item We design a VLM-based global planner that allocates frontier goals based on spatial context and semantic cues, enabling scalable multi-robot coordination without requiring task-specific training.
    \item Experiments on HM3D demonstrate that our proposed multi-robot cooperative framework significantly improves visual target navigation performance. Furthermore, we validate Co-NavGPT in real-world scenarios using two quadruped robotic platforms and achieve real-time performance in effective multi-robot navigation.

\end{itemize}

\section{RELATED WORK}


\subsection{Visual Semantic Navigation}

Visual semantic navigation is a fundamental capability for intelligent agents, inspired by human-like semantic reasoning and object search. Early classical approaches constructed metric or topological maps of the environment and planned paths to target objects accordingly. More recently, end-to-end learning methods have gained traction.
\cite{Zhu2017} introduced a reinforcement learning-based policy that encodes RGB observations and target images into a joint embedding space. Subsequent works have enhanced navigation performance through imitation learning~\cite{Ramrakhya2022}, graph-based scene representations~\cite{Lyu2022}, auxiliary perception tasks~\cite{Ye2021a}, CLIP-based vision backbones~\cite{Khandelwal2022}, and data augmentation~\cite{Maksymets2021}. However, monolithic learning policies often suffer from poor sample efficiency and generalization to unseen environments.
To address these limitations, \cite{Chaplot2020b} proposed a modular framework that separates semantic mapping, global planning, and local control, requiring learning only at the high-level planning stage. This design improves both learning efficiency and transferability. Subsequently, diverse feature representations, including topological graph~\cite{Chaplot2020a}, geometry~\cite{Chaplot2020}, and semantic~\cite{Liang2021} maps, or potential functions \cite{poni}, have been employed to train high-level policies. Zero‑shot~\cite{Majumdar2022, Al-Halah2022} and few‑shot~\cite{poni,Yu2023a} multi‑modal frameworks further enhance scene generalization.
However, single‑robot methods still face limited exploration efficiency and low fault tolerance, especially in large or complex environments. In this work, we extend visual semantic navigation to multi‑robot cooperative scenarios in unknown spaces, aiming to leverage VLMs for faster and more robust multi-robot target search.

\subsection{Multi-Robot Cooperative Navigation}

Many studies have addressed the limitations of single-robot systems by examining multi-robot cooperation across various domains, including active mapping~\cite{Ye2022}, exploration~\cite{Yu2022}, and target search~\cite{Liu2022}. Classical planning-based methods primarily focus on coordinating multiple robots for goal assignment~\cite{Puig2011}, as exemplified by challenges like the multiple traveling salesman problem. 
In exploration tasks, both multi-agent reinforcement learning~\cite{Yu2022,Yu2023} and graph-based methods~\cite{Ye2022} have been proposed to extend single-agent planners to multi-agent settings.
\cite{Liu2022} emphasized multi-agent visual semantic navigation, leveraging scene prior knowledge to locate objects within maps and subsequently formulating the navigation policy via reinforcement learning. 
While both planning-based and learning-based techniques have achieved success in multi-robot tasks, they often require real-world common-sense learning for robot assignment. In contrast, vision-language models encode rich, generalized prior knowledge, making them well-suited for multi-robot navigation. \cite{Zhang2023} employed large language models (LLMs) as planners to facilitate collaboration among agents and humans in dialogue-driven scenarios. Building on this line of work, \cite{Hong2023} incorporated domain knowledge via prompt engineering to guide multi-agent cooperation. \cite{Roco} enabled the robots to discuss and reason about task strategies using LLMs. Recent work by~\cite{shtedritski2023does} showed that structured annotations can improve VLMs interaction, motivating the adoption of mark-based visual prompting in both manipulation~\cite{liu2024moka} and navigation \cite{song2024tgs} tasks.
Building on these insights, our framework integrates a VLM as a global planner using frontier-based visual prompting. This enables contextual reasoning and robot-to-frontier assignment, resulting in more efficient exploration and search. To the best of our knowledge, VLMs have not yet been applied to these tasks.


\section{The Proposed Method}

\begin{figure*}[htbp]
    \centering
    \includegraphics[scale=0.58]{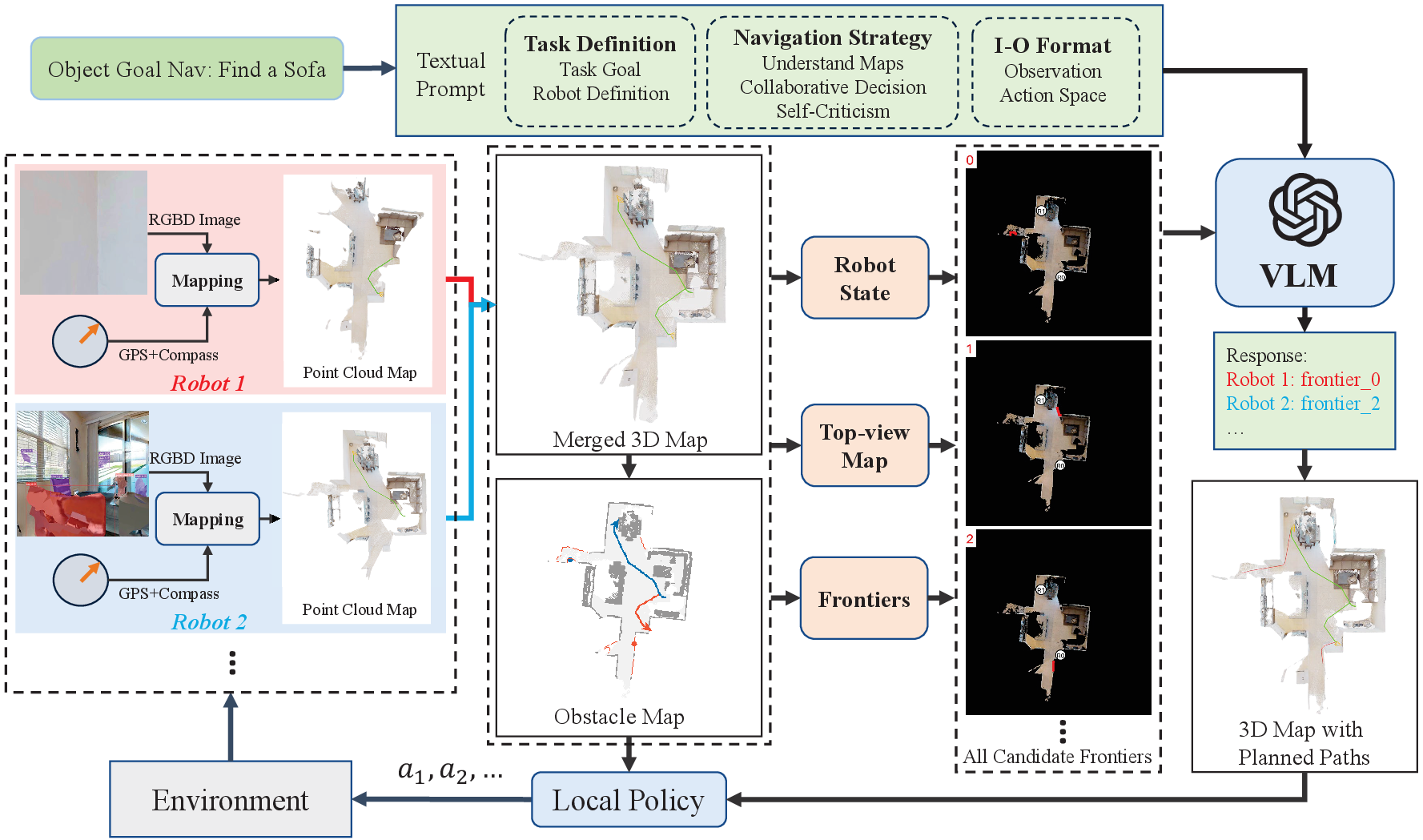}
    \caption{System architecture of the proposed multi-robot navigation framework. Each robot processes RGB-D observations to generate a local point cloud map, which is then merged into a global 3D map. The merged map, robot states, and candidate frontiers are encoded into a prompt and passed to a vision-language model, which acts as a global planner to assign frontier goals to each robot. A local policy then computes paths to the assigned frontiers based on the obstacle map, enabling coordinated exploration and target search.}
    \label{fig:system_architecture}
\end{figure*}


\subsection{Task Definition}

In the multi-robot visual target navigation task, all robots cooperate to locate an object of a specified category within unknown scenes. The  set of categories is described by $C = \left\{c_1, \dots, c_m\right\}$ and the set of scenes can be represented by $S = \left\{s_1, \dots, s_k\right\}$.
For each episode, $n$ robots $R = \left\{r^1, \dots, r^{n}\right\}$ are initialized at the same position $p_i$ within scene $s_i$, but with different initial orientations. All robots are assigned the same target category $c_i$. Thus, each episode is defined by $T_i = \left\{N, s_i, c_i, p_i\right\}$.
At each time step $t$, each robot $r^i$ receives an observation $o^i_t$ from its own perspective and performs an action $a^i_t$ simultaneously. The observation $o^i$ contains RGB-D images $I$, the location and orientation $p$ of the robot, and the object category $c$. The action space $a \in \mathcal{A}$ consists of six discrete actions: \texttt{move\_forward}, \texttt{turn\_left}, \texttt{turn\_right}, \texttt{look\_up}, \texttt{look\_down}, and \texttt{stop}. Executing \texttt{move\_forward} advances the robot by 25~cm, while the rotation actions change the robot’s orientation by 30$^\circ$ in the corresponding direction. The \texttt{stop} action is triggered when the robot is close to the target object.
An episode is considered successful if the robot issues the \texttt{stop} action while within 0.1~m of the target. Each robot is allowed a maximum of 500 time steps per episode.

\subsection{Overview}


As illustrated in Fig.~\ref{fig:system_architecture}, our framework leverages VLMs for goal selection in a multi-robot navigation setting. Each robot collects RGB-D observations to generate a local point cloud map. These individual maps are then merged into a global 3D representation based on the robots’ positions.
The merged map, along with each robot's state, a top-view map, and the set of detected frontiers, is encoded into a structured prompt and provided to the VLM. The VLM functions as a global planner, assigning distinct frontier goals to each robot based on contextual understanding.
Given these long-horizon assignments, a local policy plans collision-free paths over the obstacle map, enabling each robot to explore and search for the target object efficiently.

\subsection{Map Representation}


\subsubsection{3D Point Cloud Map}
For each robot $r^i$, given a sequence of RGB-D images $\mathcal{I}^i = \{I^i_1, \dots, I^i_t\}$ and corresponding poses $\mathcal{P}^i = \{p^i_1, \dots, p^i_t\}$, a 3D point cloud map $\mathcal{M}^i$ is constructed incrementally.
At each frame $I^i_t$, an open-vocabulary object detector $Det(\cdot)$ is first applied to identify candidate bounding boxes. A class-agnostic segmentation model $Seg(\cdot)$ then extracts segmentation masks for the detected objects. Depth information from each masked region, along with unsegmented areas, is projected into the global 3D frame using $F_{proj}(\cdot)$ based on the current pose $p^i_t$. The local map $\mathcal{M}^i$ is updated with each incoming observation.
To construct the global point cloud map $\mathcal{M}$, all local maps $\mathcal{M}^i$ are merged into a common global coordinate frame according to each robot's estimated pose:
\begin{equation}
    \mathcal{M} = \sum_{i=1}^{n} \sum_{\tau=1}^{t} F_{proj}\big(Seg(Det(I^i_\tau)), p^i_\tau\big)\,.
\end{equation}
Each semantic mask is accordingly projected onto the global semantic point cloud map. 
To further refine the map, we apply DBSCAN clustering~\cite{dbscan} to filter out noisy points.



\subsubsection{2D Exploration Map}

To support efficient exploration, we construct a 2D exploration map used for frontier extraction and path planning. The 2D map is initialized with zeros at the beginning of each episode, with the robot positioned at the center. 
All 3D point cloud data $\mathcal{M}$ is projected onto a top-down 2D grid map, which consists of two channels: an obstacle map and an explored map. Points above the floor are selected and projected onto the obstacle map, while all 3D points contribute to the explored map.
Frontiers are extracted from the boundaries of these two channels. First, the edge of the explored area is identified by detecting the largest contours in the explored map. Then, the frontier map is computed by dilating the obstacle edge and subtracting it from the explored area. 
Connected neighborhood analysis is applied to group frontier cells into clusters, and small clusters are discarded as noise. The resulting frontiers provide spatial candidates for the VLM-based planner described in the following section.

\subsection{VLM-based Multi-Robot Exploration}

After constructing the 3D point cloud and 2D exploration maps, the VLM is used to enable efficient multi-robot exploration by assigning frontiers to each robot. This is achieved via a specifically designed prompting scheme that encodes spatial and semantic context.

\subsubsection{Multi-Modal Prompting}

To handle the complexity of environments, we design both textual and visual prompts to enhance the VLM's understanding of the explored context.

\paragraph*{Visual Prompt} The visual prompt is derived from the global 3D point cloud, rendered into a colored top-view map that shares the same coordinate system and resolution as the 2D exploration map. Each robot’s position is annotated using a circle and labeled with a unique ID. Since multiple frontiers may exist at each step, we generate one visual map per candidate, each masked with a frontier and labeled with its ID in the top-left corner to assist the VLM in identification. These maps provide both semantic and geometric information, enabling the VLM to reason over spatial configurations.
An example of the visual prompt is shown in Fig.~\ref{fig:system_architecture}. 

\paragraph*{Textual Prompt} The textual prompt encodes the task objective, environment assumptions, and expected output format. It is designed to clarify the task structure and facilitate effective frontier assignment. The prompt includes: (i) a task description, (ii) contextual background, (iii) reasoning requirements, and (iv) a standardized input-output format. An example is shown below:

\begin{tcolorbox}
    \fontsize{8}{8}\selectfont
    \textbf{Task}: Locate the given target\\
    
    \textbf{Context}:\\
    We have multiple robots. Each robot perceives the environment and can navigate to explore unknown areas.\\
    The global top-view map shows the positions of each robot, candidate frontiers, and their IDs.\\
    
    \textbf{Requirements}:\\
    Understand: the scene layout, the robots' state, and the frontiers.\\
    Analyze: collaborative exploration and efficient target searching.\\
    Decide: a frontier assignment policy such that each robot moves to an optimal frontier.\\
    Justify: Reconsider the decision with a concise explanation.\\
    
    \textbf{Input}: Multiple top-view maps, each containing one candidate frontier. A target object category is provided.\\
    \textbf{Output}: A JSON object indicating the frontier IDs assigned to each robot.\\
\end{tcolorbox}

Given the above, the textual and visual prompts inject structured commonsense into the VLM, improving its ability to generate accurate and explainable frontier assignments. Once the prompt is complete, it is passed to the VLM to obtain the final assignment for each robot.

\subsubsection{Decision Making}

As noted in prior work~\cite{Huang2022b, Song2022}, while vision-language models are effective at generating high-level plans, they are less reliable for fine-grained low-level control. To address this, we use the VLM solely for global goal assignment.
At each global step, an updated point cloud map is used to construct a visual prompt that captures both environmental and robot-specific information. The VLM then assigns a frontier waypoint to each robot, as illustrated in Fig.~\ref{fig:system_architecture}. The frontier assignment is guided by two criteria: (i) collaborative exploration of unknown regions, and (ii) semantic relevance to the target object. This encourages the selection of frontiers that not only expand the map coverage but are also more likely to be near the target based on the top-view semantic layout.
If no viable frontiers are available, random points within the explored space are selected as fallback goals. The global planner updates these long-term goals every 25 local steps.

\subsubsection{Local Policy}

Once a long-term goal is assigned, each robot plans a path using the Fast Marching Method (FMM)~\cite{Sethian1996} from its current position to the goal. A short-range local goal is then selected along this path. The final action $a_t \in \mathcal{A}$ is computed to reach this local goal, taking into account obstacles and robot dynamics.
At each step, the robot updates its local map and recalculates the local goal based on new sensor observations. This local policy compensates for the limitations of VLMs in fine-grained decision-making, ensuring smooth and efficient navigation.

\section{EXPERIMENTS}

\begin{figure*}[htbp]
	\centering
	\subfloat[step 25]
	{
		\includegraphics[width=4.2cm]{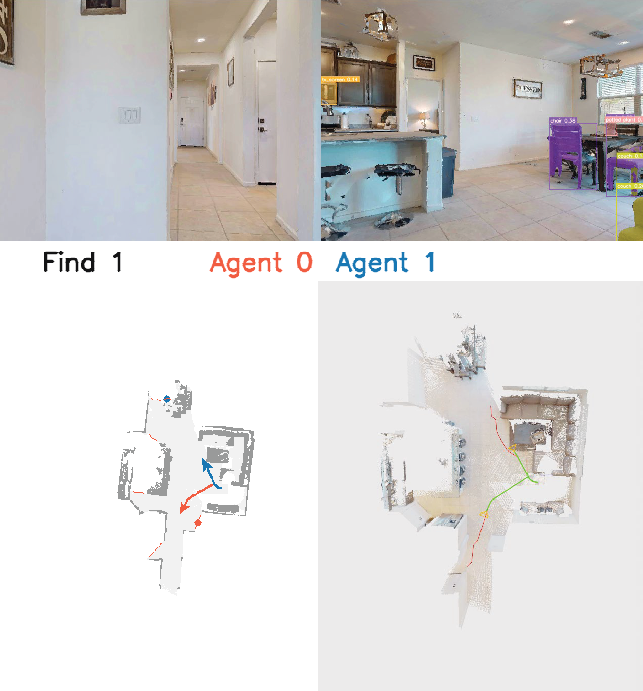}
	}
	\subfloat[step 50]
	{
		\includegraphics[width=4.26cm]{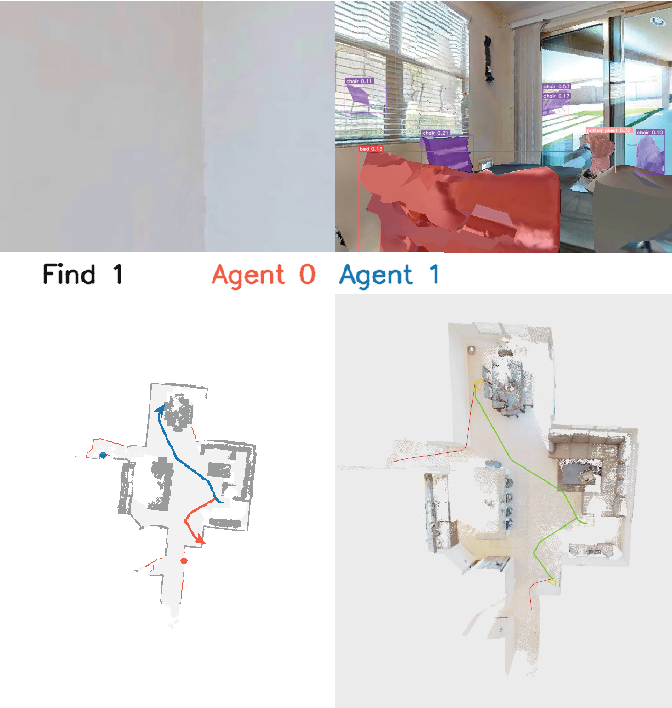}
	}
	\subfloat[step 75]
	{
		\includegraphics[width=4.2cm]{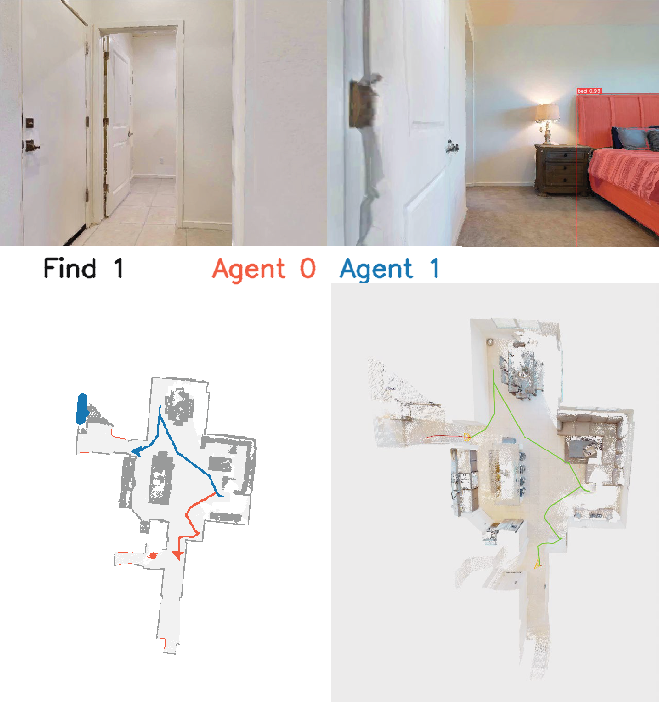}
	}
	\subfloat[step 85]
	{
		\includegraphics[width=4.2cm]{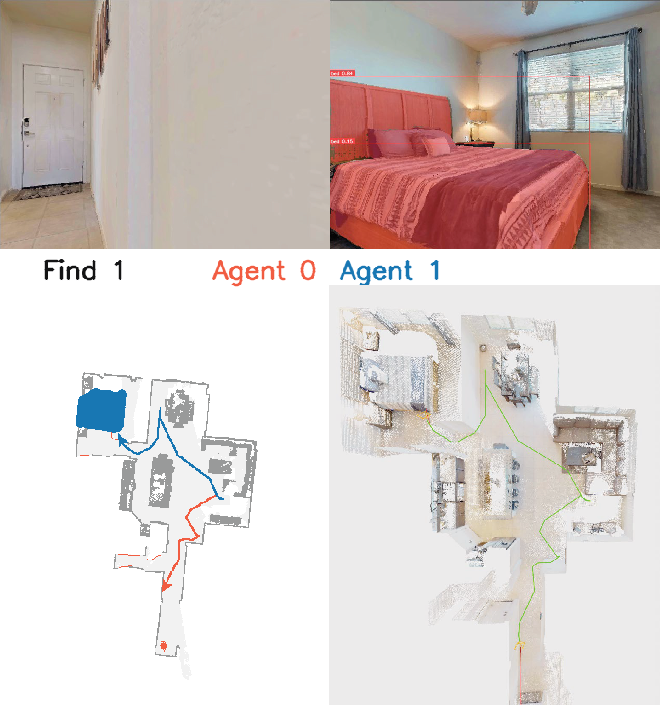}
	}
	\caption{Visualization of the visual target navigation process in the Habitat simulator using two robots to locate a bed. The top row shows first-person RGB observations from both robots (Agent 0 in red, Agent 1 in blue) at different time steps. The bottom row presents the corresponding 2D exploration map and 3D point cloud map. The red and blue lines represent the respective paths of Agent 0 and Agent 1. Red and blue dots denote frontier goals assigned by the vision-language model.}
	\label{fig:sim_episode}
	\vspace{-0.4cm}
\end{figure*}

In this section, we compare our method with other multi-robot map-based baselines in the simulation to evaluate the performance of our framework. Additionally, we apply our process in two real-world robot platforms to validate its practicality for multi-robot navigational tasks.

\subsection{Simulation Experiment}
\subsubsection{Dataset}
We conduct experiments using the HM3D\_v0.2 dataset~\cite{Ramakrishnan2021a}, which features high-resolution, photorealistic 3D reconstructions of real-world environments. Compared to earlier datasets~\cite{Zhu2017,Gibson}, HM3D offers larger and more complex scenes. Following the standard protocol, we use 36 validation scenes containing 1,000 episodes with semantic object annotations. We select six goal categories, following~\cite{Chaplot2020b}: \textit{chair, sofa, plant, bed, toilet}, and \textit{TV}.

\subsubsection{Settings}
\label{sec:settings}
All experiments are conducted using the Habitat simulator~\cite{Savva2019}. Each robot receives $480 \times 640$ RGB-D observations, odometry information, and a goal category encoded as an integer. We use YOLOv8~\cite{yolov8_ultralytics} for open-vocabulary detection $Det(\cdot)$ and Mobile-SAM~\cite{mobile_sam} for class-agnostic segmentation $Seg(\cdot)$. The 2D exploration map covers a $24 \times 24$\,m area with a resolution of 0.05\,m.
For global planning, we employ GPT-4o~\cite{openai2024gpt4o} via the OpenAI API as the VLM. 
A visual prompt is generated at each global planning step to represent all robots and the environment state, and the VLM outputs a JSON to assign frontier goals.
Our implementation is based on publicly available codes from~\cite{vln-game}, using two robots per episode, both starting from the same position with different initial orientations. The 3D point cloud map and robot state are visualized using Open3D~\cite{open3d}.

\subsubsection{Evaluation Metrics}

We follow \cite{Anderson2018} to evaluate our method using the Success Rate (SR), Success weighted by the Path Length (SPL), and Distance to Goal (DTG) for multi-robot tasks. SR is defined as $\frac{1}{N} \sum_{i=1}^{N} S_{i}$, and SPL is defined as $\frac{1}{N} \sum_{i=1}^{N} S_{i} \frac{l_{i}}{\max \left(l_{i}, p_{i}\right)}$, where $N$ is the number of episodes, $S_{i} = 1$ is considered successful if any robot successfully locates the target; otherwise, the episode is deemed a failure. $l_{i}$ denotes the shortest trajectory length from the start position to one of the success positions, $p_{i}$ stands for the shortest robot's trajectory length in the current episode $i$. Lastly, DTG represents the minimal distance between the robots and the target goal when the episode ends.

\subsubsection{Baselines}

To evaluate the navigation performance of our method, we compare it against several baselines. All baselines share the same framework for map construction using object detection; however, they differ in the global policy used for assigning robots to frontiers. After frontier selection, all robots follow the same local planning policy to execute actions.

\begin{itemize}

	\item \textbf{Greedy\cite{Visser2013}: }Frontiers are assigned to robots in a greedy manner. Each robot selects the nearest unassigned frontier as its goal.
	\item \textbf{Cost-Utility~\cite{Julia2012}:} Each frontier cell $f \in F$ is scored using a cost-utility function:
    \begin{equation}
        S^{CU}(f) = U(f) - \lambda_{CU} C(f)\,,
    \end{equation}
    where $U(f)$ denotes the utility based on frontier size, and $C(f)$ represents the distance from the robot to the frontier. The parameter $\lambda_{CU}$ balances the trade-off between utility and cost. Each robot selects the frontier with the highest $S^{CU}(f)$ score.
	\item \textbf{Random Sampling:} Inspired by~\cite{Yu2023a}, this method randomly samples long-term goals from the map. This baseline reflects the potential of stochastic goal selection in complex and unstructured environments.
    \item \textbf{Multi-SemExp~\cite{Chaplot2020b}:} We extend the SemExp framework to a multi-robot setting. Two robots jointly explore the environment, and the task is considered successful if either robot finds the target object.

\end{itemize}

\renewcommand\arraystretch{1.5}
\begin{table}[t]
	\centering
	\fontsize{9}{9}\selectfont
	\begin{threeparttable}
		\caption{Results of Comparative Study.}
		\label{tab:performance_comparison}
		\setlength{\tabcolsep}{3.5mm}{}
		{
			\begin{tabular}{cccc}
				\toprule
				{Method}                        & SR$\uparrow$ & SPL$\uparrow$  & DTG$\downarrow$		  \cr
				\midrule
				Greedy\cite{Visser2013}         & 0.611       & 0.328       & 2.239      \cr
				Cost-Utility\cite{Julia2012}    & 0.625       & 0.323       & 2.030         \cr
				Random Sample on Map            & 0.636       & 0.336       & 2.048         \cr
				Multi-SemExp\cite{Chaplot2020b} & 0.612       & 0.327       & 2.234         \cr
				Co-NavGPT (Ours)                & {\bf 0.666} & {\bf 0.368} & {\bf 1.684}   \cr
				\bottomrule
			\end{tabular}
		}

	\end{threeparttable}
\end{table}

\subsubsection{Results and Discussion}

The quantitative results are summarized in Table~\ref{tab:performance_comparison}. Among traditional baselines, the Cost-Utility method outperforms the Greedy strategy, demonstrating the benefit of jointly considering frontier size and travel cost during goal assignment.
Interestingly, the random sampling strategy achieves even better performance than Cost-Utility. This suggests that long-range random exploration can help robots quickly spread out across large, unknown environments and cover more ground early.
The Multi-SemExp baseline does not perform as well, likely due to limitations in its global policy, which was trained on small-scale local maps~\cite{Chaplot2020b} and does not generalize well to larger multi-robot settings.
Our proposed method, Co-NavGPT, consistently outperforms all baselines across all three metrics (SR, SPL, and DTG). These results demonstrate the effectiveness of the VLM-based reasoning in assigning semantically meaningful and spatially efficient goals for collaborative exploration.
An example trajectory of a successful episode in which the target is a bed is illustrated in Fig.~\ref{fig:sim_episode}. The sequence demonstrates how robots collaboratively explore different regions to efficiently locate the target.

\subsubsection{Ablation Study}

To evaluate the contribution of individual components in our framework, we conduct ablation studies on the HM3D dataset using the following variants:

\begin{itemize}
    \item \textbf{Single:} A single-robot configuration guided by a VLM-based global planning.
    \item \textbf{Mini.:} Replacing GPT-4o~\cite{openai2024gpt4o} with the lighter GPT-4o-mini~\cite{openai2024gpt4omini} to assess the impact of the VLM scale.
    \item \textbf{Obs.:} Replacing the top-view semantic map with an obstacle-only map, while maintaining identical visual prompting format.
\end{itemize}

The results, shown in Table~\ref{tab:ablation_study}, demonstrate that our full model achieves the best performance in both success rate and SPL. Replacing the top-view map with the obstacle map (\textbf{Obs.}) leads to a drop in success rate, indicating the importance of semantic context (e.g., color and layout) in the VLM-based decision making.
Substituting GPT-4o with GPT-4o-mini (\textbf{Mini.}) reduces performance across both metrics, suggesting that the reasoning capability of larger VLMs contributes significantly to task effectiveness.
Finally, the single-robot configuration (\textbf{Single}) yields the lowest performance, highlighting the robustness and scalability benefits of the multi-robot framework.

\renewcommand\arraystretch{1.4}
\begin{table}[t]
	\centering
	\fontsize{9}{9}\selectfont
	\begin{threeparttable}
		\caption{Results of Ablation Study in HM3D.}
		\label{tab:ablation_study}
		\setlength{\tabcolsep}{1mm}{}
		{
			\begin{tabular}{ccccccc}
				\toprule
				Ablation        &NUM             & Visual Prompt         & VLM              & Success$\uparrow$          & SPL$\uparrow $            \cr 
				\midrule
				Single          &1               & Top-view Map          & GPT4o             & 0.579                     & 0.211                     \cr 
				Mini.           &2               & Top-view Map          & GPT4o-mini        & 0.641                     & 0.307                     \cr 
				Obs.            &2               & Obstacle Map          & GPT4o             & 0.645                     & 0.363                     \cr 
				Ours            &2               & Top-view Map          & GPT4o             & 0.666                     & 0.368                     \cr 
				\bottomrule
			\end{tabular}
		}

	\end{threeparttable}
\end{table}

\subsubsection{Failure Cases}
Although our framework achieves high success rates across a wide range of scenes, we observe several failure cases that highlight its current limitations and opportunities for future improvement.
In the experiments in the HM3D dataset, approximately 14\% of failures are attributed to exploration issues, while around 20\% stem from object detection errors.
Exploration-related failures are primarily caused by situations where robots become trapped in incomplete or poorly reconstructed regions of the scene (e.g., topological holes), or where the target object is either absent, undetectable, or located on a different floor. Multi-floor navigation remains a challenge under the current setup.
Detection-related failures mainly result from misclassifications or missed detections, which serve as the primary detection backbone. In some cases, robots stop prematurely at positions distant from the actual target due to incorrect detections.
Addressing these limitations, particularly in robust multi-level planning and more reliable open-vocabulary object detection, remains an important direction for future work.

\subsection{Real-World Experiment}

\begin{figure}[t]
	\centering
	\subfloat[Target: A sink.]
	{
		\includegraphics[width=8.4cm]{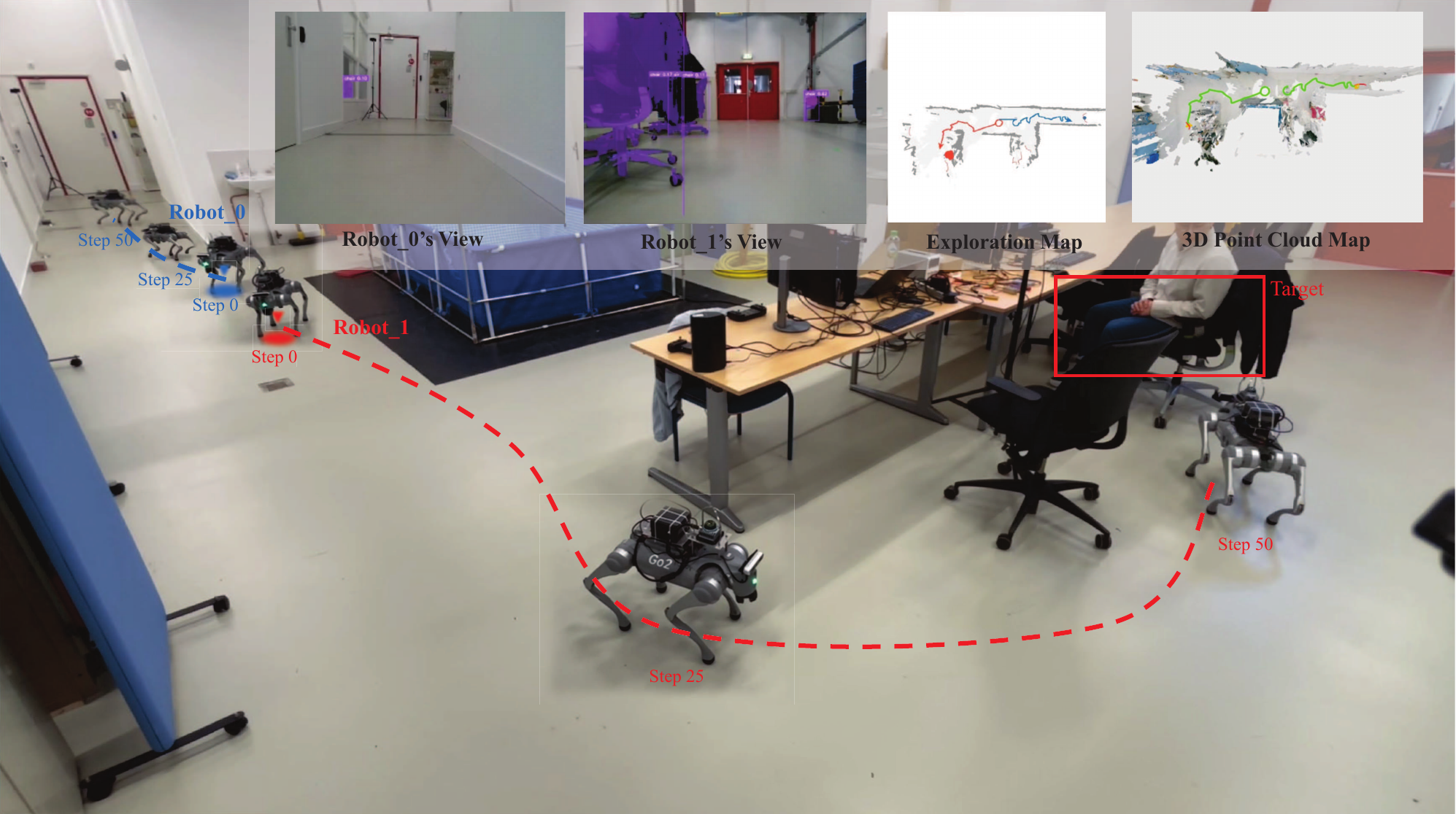}
	}
    
	\subfloat[Target: A person.]
	{
		\includegraphics[width=8.4cm]{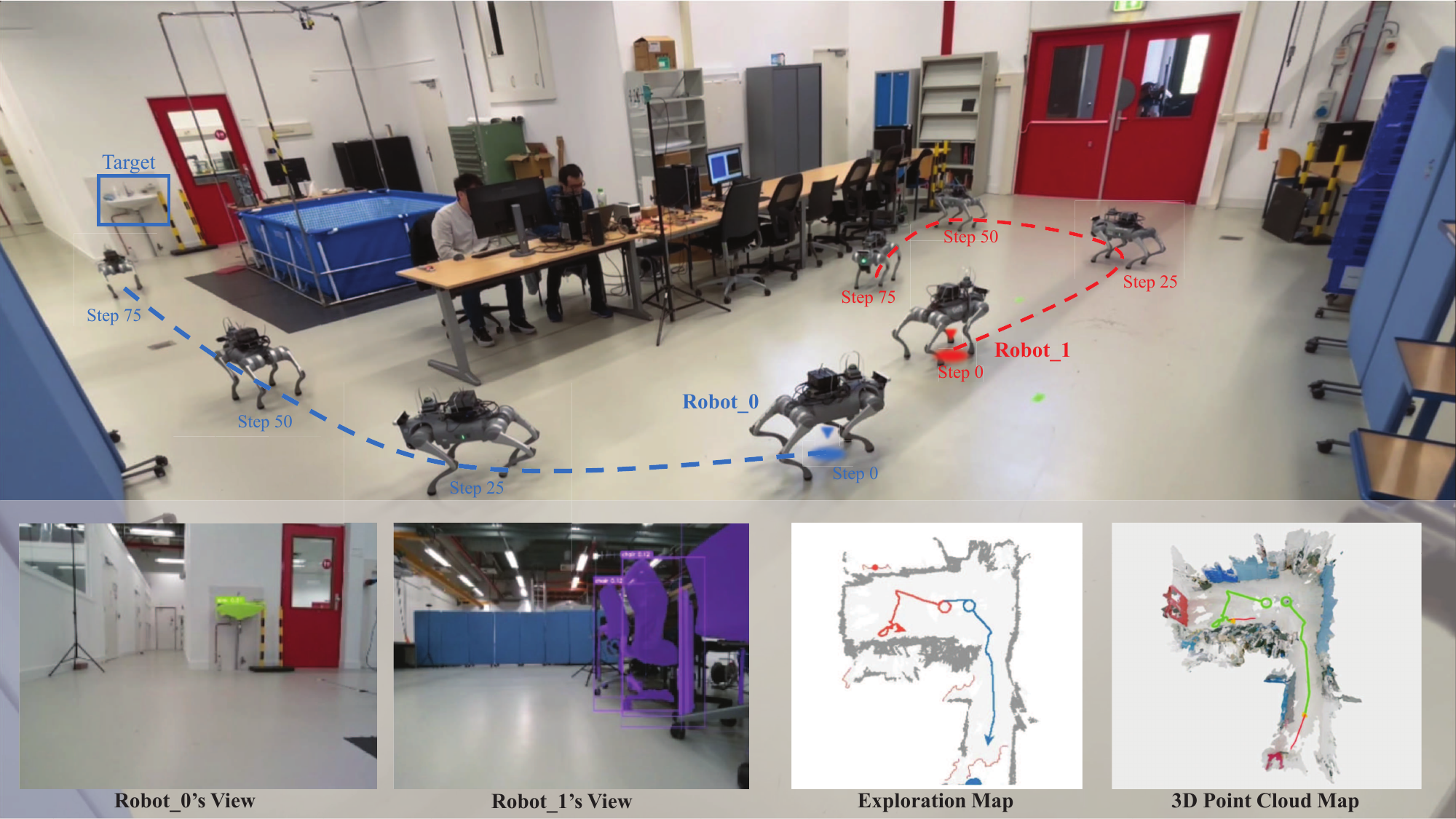}
	}

        \subfloat[Target: A chair.]
	{
		\includegraphics[width=8.4cm]{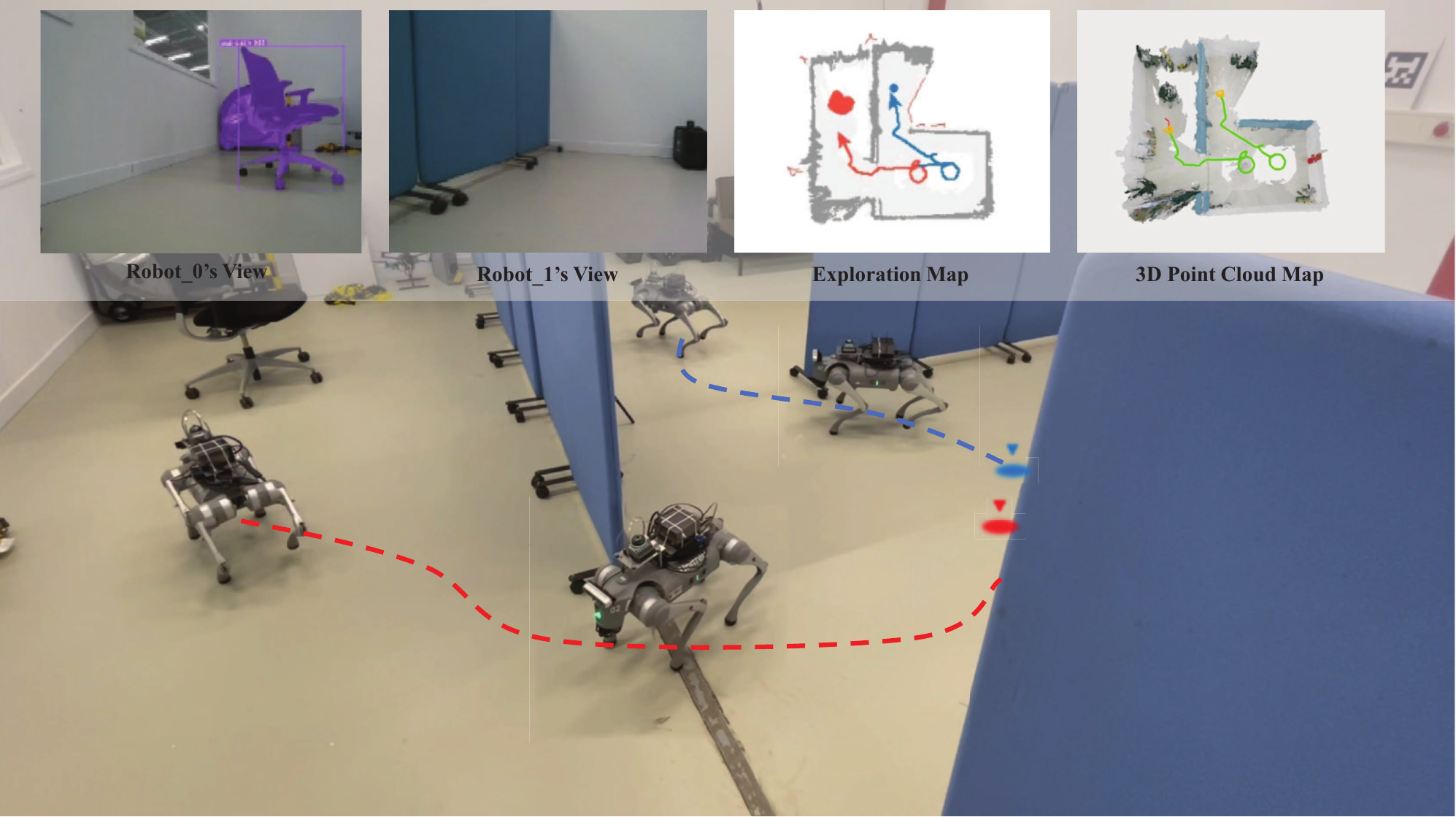}
	}
    
	\caption{Real-world multi-robot visual target navigation. Each scene shows the first-person RGB images, the exploration map, and the 3D point cloud map. In (a), the robots search for a sink; (b) search for a person; (c) search for a chair. Red and blue lines denote the trajectories of the two robots.}
	\label{fig:real}
	\vspace{-0.4cm}
\end{figure}

We implement our framework for two Unitree Go2 quadruped robots, each equipped with a RealSense D455 RGB-D camera and a Livox Mid-360 LiDAR. The multi-agent system is deployed in a real-world, previously unseen lab environment. The navigation tasks include locating objects, namely, a chair, a sink, and a person.
To facilitate the real-world deployment, we configure the sensors and modules according to the settings as \ref{sec:settings} in the simulation. The RGB-D camera is adjusted to match the resolution and depth range as in the simulation. Due to sensor noise caused by lighting and hardware variations, we apply DBSCAN~\cite{dbscan} to cluster dense point regions and filter outliers in the point cloud.
We exploit the onboard Livox Mid-360 LiDAR with the built-in IMU and employ FAST-LIO2 for localization~\cite{fastlio2}, and conduct camera-to-LiDAR extrinsic calibration~\cite{calibration} to align depth data into the global frame. Before exploration begins, we apply Generalized-ICP~\cite{gicp} to compute initial pose alignment between the two robots, setting Robot 0’s initial frame as the global reference.
Once the map is constructed from RGB-D observations, frontiers are extracted and assigned by the vision-language model. Based on the map and robot positions, our framework generates waypoints and corresponding discrete navigation actions.
Figure~\ref{fig:real} demonstrates three cases, where the proposed Co-NavGPT enabled cooperative navigation of different targets using the two quadruped robots. In each scenario, the VLM successfully assigns complementary frontiers to the robots, enabling cooperative exploration and efficient target search. The entire system runs in real time at approximately 5~FPS, which is sufficient for multi-robot navigation tasks in indoor environments.

All perception and planning modules, including YOLOv8 \cite{yolov8_ultralytics}, Mobile-SAM~\cite{mobile_sam}, and our VLM-based planner, are deployed on a workstation equipped with an NVIDIA RTX 4090 GPU. Sensor processing, LiDAR-based localization, and low-level control modules run onboard the robots.
Our framework is hardware-agnostic and can be flexibly deployed, as it only requires RGB-D observations and pose estimates as input, and outputs discrete navigation actions.

\section{Conclusions}

We presented Co-NavGPT, a novel framework that leverages vision-language models for multi-robot cooperative visual target navigation. By encoding scene-level information into structured visual prompts, a VLM functions as a global planner, enabling efficient frontier assignment for collaborative exploration and object search.
Extensive experiments in simulation demonstrate that our approach significantly outperforms existing multi-robot baselines in terms of success rate and path efficiency, without relying on any task-specific learning. Real-world experiments also validate its practicality for multi-robot navigational tasks.
These results highlight the strong potential of VLMs in coordinating complex multi-agent behaviors. Future work includes investigating tighter integration between VLMs and embodied agents in 3D environments, particularly toward interactive decision-making, dynamic replanning, and closed-loop real-time control.



\bibliographystyle{ieeetr}
\bibliography{bib/library.bib}

\end{document}